\documentclass[11pt,twoside]{article}
\usepackage[latin9]{inputenc}
\usepackage{array}
\usepackage{textcomp}
\usepackage{graphicx}
\usepackage{url}

\makeatletter

\DeclareRobustCommand{\greektext}{%
  \fontencoding{LGR}\selectfont\def\encodingdefault{LGR}}
\DeclareRobustCommand{\textgreek}[1]{\leavevmode{\greektext #1}}
\DeclareFontEncoding{LGR}{}{}
\DeclareTextSymbol{\~}{LGR}{126}
\providecommand{\tabularnewline}{\\}

\usepackage{jmlr2e}

\ShortHeadings{Deep Convolutional Networks for Point of Care Diagnostics}{Deep Convolutional Networks for Point of Care Diagnostics}

\firstpageno{1}

\makeatother

\begin{document}

\title{Deep Convolutional Neural Networks for \\Microscopy-Based Point
of Care Diagnostics}

\author{\name John A. Quinn$^\dagger$ \email jquinn@cit.ac.ug
\AND
\name Rose Nakasi$^\dagger$ \email g.nakasi.rose@gmail.com
\AND
\name Pius K. B. Mugagga$^\dagger$ \email piuskavz@gmail.com
\AND
\name Patrick Byanyima$^\ddag$ \email byanyimap@yahoo.com
\AND
\name William Lubega$^\dagger$ \email drlubega.william@gmail.com
\AND
\name Alfred Andama$^\ddag$ \email andama.alf@gmail.com\\
\addr $^\dagger$College of Computing and Information Sciences, Makerere University, Uganda\\
$^\ddag$College of Health Sciences, Makererere University, Uganda\\
}

\maketitle
\begin{abstract}
Point of care diagnostics using microscopy and computer vision methods
have been applied to a number of practical problems, and are particularly
relevant to low-income, high disease burden areas. However, this is
subject to the limitations in sensitivity and specificity of the computer
vision methods used. In general, deep learning has recently revolutionised
the field of computer vision, in some cases surpassing human performance
for other object recognition tasks. In this paper, we evaluate the
performance of deep convolutional neural networks on three different
microscopy tasks: diagnosis of malaria in thick blood smears, tuberculosis
in sputum samples, and intestinal parasite eggs in stool samples.
In all cases accuracy is very high and substantially better than an
alternative approach more representative of traditional medical imaging
techniques.
\end{abstract}

\section{Introduction}

Conventional light microscopy remains the standard method of diagnosis
for a number of conditions, such as malaria. Microscopy is particularly
well adapted to low-resource, high disease burden areas, being both
simple and versatile; even for diagnostic tasks for which newer technologies
are available, e.g. based on flow cytometry or molecular biology,
the cost of specialised equipment may render it impractical in such
places. 

In contrast to alternatives such as rapid diagnostic tests, however,
microscopy-based diagnosis does depend on the availability of skilled
technicians, of which there is a critical shortage. A nationwide study
in Ghana, for example, found 1.72 microscopes per 100,000 population,
but only 0.85 trained laboratory staff per 100,000 population \citep{2004ghanadiagnostics}.
As a result, diagnoses are often made on the basis of clinical signs
and symptoms alone, which is error-prone and leads to higher mortality,
drug resistance, and the economic burden of buying unnecessary drugs
\citep{2006laboratoriesafricabarrier}.

There is therefore need for alternatives which help to provide the
access to quality diagnosis that is currently routinely unavailable.
In this work, we focus on the development of point-of-care (POC) diagnostics
which utilise two relatively common resources: microscopes and smartphones.
Smartphones are widely owned across the developing world, and have
the capacity to capture, process and transmit images. Given the appropriate
hardware to couple the two, this setup has enormous potential for
remote and automated diagnosis \citep{2015smartphonediagnostics}.
In principle, any microscopical assessment can be automated with computer
vision methods, within the limits of camera optics and the accuracy
of image analysis methods.

The field of computer vision has been significantly advanced recently
by the emergence of deep learning methods, to the extent that some
object recognition tasks can now be automated with accuracy surpassing
human capability \citep{2015surpassingimagenet}. Rather than relying
on the extraction of image features hand-engineered for a particular
task, a standard approach in medical imaging, such methods learn effective
representations of input images automatically, with successive layers
in the model representing increasingly complex patterns. Recent work has shown the efficacy of these techniques on the detection of malaria \citep{sanchez2016malaria} and intestinal parasites \citep{peixinho2015diagnosis} detection, as well as other microscopy diagnosis tasks including mitosis \citep{cirecsan2013mitosis} and C. elegans embryo \citep{ning2005embryo} detection. 
This paper demonstrates the application of deep learning to microscopy-based POC diagnostics, with particular focus on the end-to-end application of these methods in a resource-constrained environment, using images captured by a low cost smartphone microscope adapter developed for this study. We provide experimental results for three diagnostic tests: malaria (in blood smear samples), tuberculosis (in sputum samples) and intestinal parasites (in stool samples).

The structure of the paper is as follows: we briefly provide background
on the three diagnostic tasks addressed in this work, and in Section
\ref{sec:Microscopy-Image-Capture} describe the data collection setup,
including the use of 3D-printed adapters to couple the phone to the
microscope. In Section \ref{sec:Methods} we describe the computer
vision methodology, using convolutional neural networks to learn to
distinguish the characteristics of pathogens in different types of
sample image. Section \ref{sec:Results} gives experimental results,
and Section \ref{sec:Discussion-and-Related} concludes and discusses
links to previous work.

Full code, data and 3D hardware models to recreate the experiments
in this paper are available online at \url{http://air.ug/microscopy}.

\subsection{Background}

\textbf{Malaria} is caused by parasites of the genus \emph{plasmodium},
and the gold standard diagnosis is the microscopical examination of
stained blood smear samples to identify these parasites. In thin films
the red blood cells are fixed so the morphology of the parasitized
cells can be seen. Species identification can be made, based on the
size and shape of the various stages of the parasite and the presence
of stippling (i.e. bright red dots) and ambriation (i.e. ragged ends).
However, malaria parasites may be missed on a thin blood film when
there is a low parasitemia. Therefore, examination of a thick blood
smear, which is 20-40 times more sensitive than thin smears for screening
of plasmodium parasites, with a detection limit of 10-50 trophozoites/\textgreek{m}L
is recommended \citep{2009malariadiagnosisreview}. 

\textbf{Tuberculosis} (TB) is ranked as a leading cause of death from
an infectious disease worldwide. Rapid screening of TB is possible,
but in rural, developing-world settings is still difficult because
current diagnostic methods are expensive, time consuming (several
days to weeks), and require specialized equipment that are not readily
available in low resource, high-burden TB areas. Currently, diagnosis
of TB relies mainly on demonstration of the presence of mycobacteria
in clinical specimens by serial sputum tests; smear microscopy, Gene
Xpert\textregistered{} MTB/RIF assay and culture. Acid-fast stains
such as the Ziehl-Neelsen (ZN) stain or the fluorescent auramine-rhodamine
stain are recommended for mycobacteria. The Ziehl-Neelsen stain forms
a complex between the mycolic acids of the mycobacterial cell wall
and the dye (e.g., carbol fuchsin) which makes the mycobacteria resistant
to decolorization by acid-alcohols unlike non-acid fast bacteria.
At least 300 fields should be examined under high power (1000x) when
using a carbol-fuchsin stained smear and light microscopy to look for
red/pink rods/bacilli against a blue background \citep{2015tbdiagnosis}. 

\textbf{Intestinal parasites, }and in particular helminths, like any
other type of parasitic organisms have a consistent external and internal
morphology throughout the different stages of development that is
egg, larva and adult stages. The typical diagnostic procedure involves
placing a drop of normal saline (0.85\% Nacl) on a clean labeled slide
and using an applicator stick a small portion of feces (approximately
2 mg, about the size of a match head) is added to the drop of saline
and emulsified to form suspension. A cover slip is gently applied
on the preparation so as not to trap air bubbles. The preparation
should be uniform: not too thick that fecal debris can obscure organisms,
nor too thin that blank spaces are present. The preparation is examined
with 10x objective to cover the entire coverslip. For hookworm, for
example, this can be done accurately when a sample contains a minimum
of 300-500 eggs per gram \citep{1949hookworm}. One challenge in identifying
helminth eggs is distinguishing them from fecal impurities such as
undigested fat, which can resemble them in appearance.

\section{Microscopy Image Capture\label{sec:Microscopy-Image-Capture}}

We briefly describe the experimental setup used to collect data for
the study, and for system prototype testing.

\subsubsection*{Hardware design}

In order to deploy computer vision methods for decision support and 
automated diagnostics, a suitable deployment platform is needed. While
there are a range of digital microscopes and imaging solutions, they
tend to be costly, or limited to particular models of microscope,
and therefore not well suited to this task. Furthermore, we found
that existing low cost smartphone adapters were awkward to use in
practice. These were very sensitive to movement (even touching the
phone's screen to take an image could cause it to lose alignment). 

\begin{figure}[t]
\centering{}\includegraphics[width=0.32\textwidth]{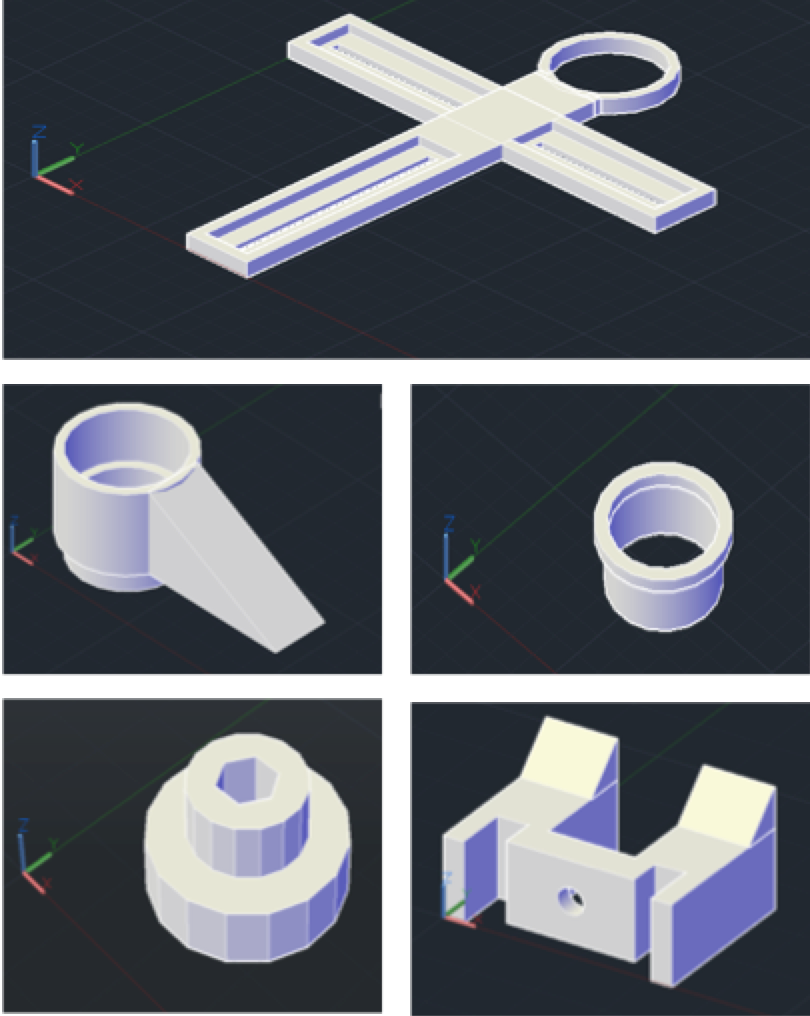}\hspace{0.3cm}\includegraphics[width=0.3\textwidth]{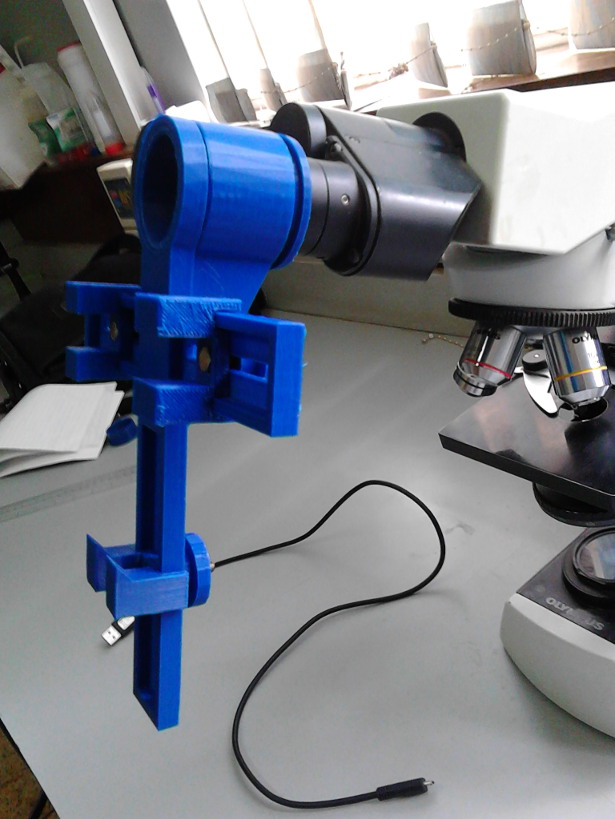}\hspace{0.3cm}\includegraphics[width=0.3\textwidth]{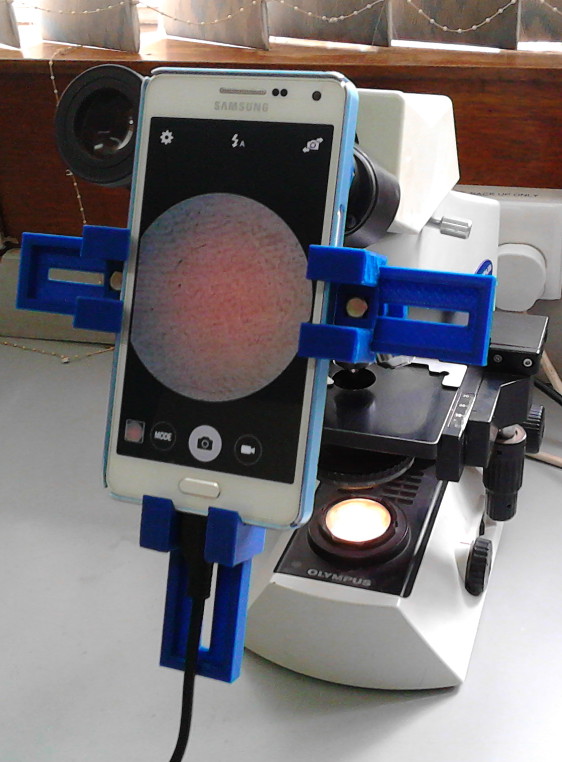}\caption{Microscope smartphone adapter: design of components (left), 3D-printed
adapter mounted on microscope (center), smartphone inserted into adapter
(right).}
\label{fig:adapter}
\end{figure}

Our improved design is shown in Figure \ref{fig:adapter}. The attachment
mechanism couples with the microscope eyepiece (ocular). The adjustment
mechanism is designed to enable a user correctly align a smartphone
of nearly any model with the focal point of the eyepiece. The mechanism
is achieved by using sliders and side-holders, which enable positioning
of the phone. The locking mechanism is used to lock the smartphone
in position once the appropriate alignment has been made. Once the
adapter is locked in alignment, the user can easily slide the smartphone
in and out of the adapter at any time without compromising the pre-set
alignment.

\begin{figure}

\centering{}\includegraphics[height=6cm]{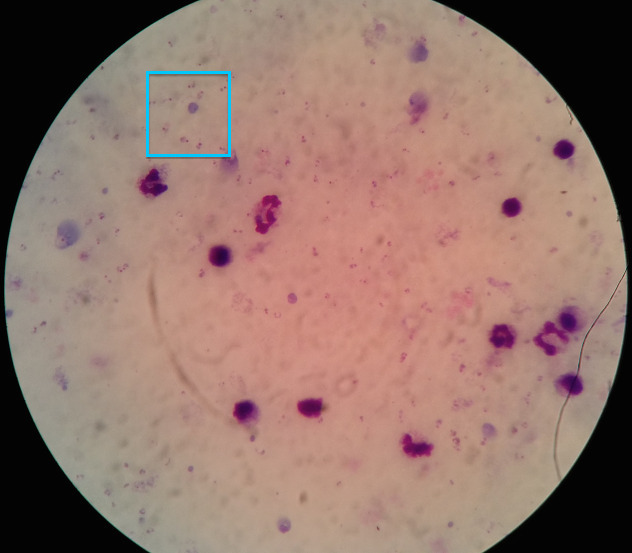}\hspace{0.3cm}\includegraphics[height=6cm]{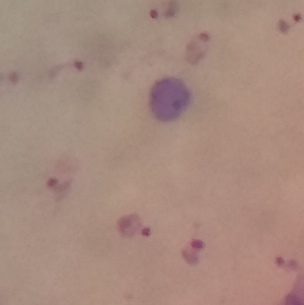}\caption{Image of thick blood smear with Giemsa stain, taken with the apparatus
shown in Fig \ref{fig:adapter}. Detail on right shows several \textit{P.
falciparum} clearly visible.}
\end{figure}

\subsubsection*{Imaging of samples}

Using the setup above, malaria images were taken from thick blood smears and stained using
Field stain at x1000 magnification. The TB images were made from fresh sputum and stained
using ZN (Ziehl Neelsen) stain. These were examined under x1000 magnification.
Finally the intestinal parasites images were captured from slides
of a wet preparation, i.e. a portion of stool sample mixed in a drop
of normal saline and examined under x400 magnification.

\subsubsection*{Image annotation}

Laboratory experts then provided input on the locations of objects
within images, which was recorded using annotation software developed
for the task. The experts identified bounding boxes around each object
of interest in every image. In thick blood smear images, plasmodium
were annotated (7245 objects in 1182 images); in sputum samples,
tuberculosis bacilli were annotated (3734 objects in 928 images),
and in stool samples, the eggs of hookworm, Taenia and Hymenolepsis
nana were annotated (162 objects in 1217 images).

\section{Methods\label{sec:Methods}}

In this section, we describe the process for training a deep learning
model from the annotated images, and then apply this model to test
images in order to detect pathogens.

\subsubsection*{Generation of training/testing set}

Each image collected was downsampled and then split up into overlapping
patches, with the downsampling factor and patch size determined by
the type of pathogen to be recognised in each case. Visual inspection
was used in each case, so that patches were large enough to contain
all pathogen objects with a small margin, and detailed enough to clearly
identify objects by eye, but without excessive detail that would add
unnecessary processing burden. Positive patches (i.e. those containing
plasmodium, bacilli or parasite eggs respectively) were taken centered
on bounding boxes in the annotation. Negative patches (i.e. with absence
of any of these pathogens, though possibly with other types of objects
such as staining artifacts, blood cells or impurities) were taken
from random locations in each images not intersecting with any annotated
bounding boxes. 

Because most of each image does not contain pathogen objects, the
potential number of negative patches is usually disproportionately
large compared to the number of positive examples. Two measures were
taken to make the training and testing sets more balanced. First,
negative patches were randomly discarded so that there was at most
100 times the number of positive patches. Second, new positive patches
were created by applying all combinations of rotating and flipping,
giving 7 extra positive examples for each original.

\subsubsection*{Specification and training of convolutional neural networks}

Convolutional neural networks (CNNs) are a form of neural network
particularly well adapted to the processing of images. Rather than
densely connected layers in traditional networks such as the classic
multi-layer perceptron structure, the sharing of weights between many
locally receptive fields means that the number of parameters is relatively
low. These locally receptive fields are convolutions over a small
region of the input. Different convolution filters respond to the
presence of different types of patterns in the input; in the initial
layers, these responses may be to edges, blobs or colors, whereas
in higher levels, the responses can be to higher-level, more complex
patterns. CNNs generally include some combination of the following
types of layers.
\begin{itemize}
\item Convolution layers are computed by taking a sliding window (the receptive
field) across the input, calculating the response function at each
location for each filter. Multiple filters capture different types
of patterns. 
\item Pooling layers reduce the size of the input, merging neighbouring
elements e.g by taking the maximum. This reduces the number of parameters,
and hence the amount of computation, as each pooling is done.
\item Fully connected layers have connections from all activations in the
previous layer to all outputs. This is equivalent to a convolutional
layer with one filter, the same size as the input. A fully connected
layer is typically used as the last layer in a CNN, with the output
having one element per class label.
\end{itemize}
In this work, we used networks with four hidden layers:
\begin{enumerate}
\item Convolution layer: 7 filters of size $3\times3$.
\item Pooling layer: max-pooling, factor 2.
\item Convolution layer: 12 filters of size $2\times2$.
\item Fully connected layer, with 500 hidden units.
\end{enumerate}
A particular configuration of layers defines a loss function for the
CNN. This loss function can be computed for a training data set, and
minised using optimisation methods. We used the Lasagne\footnote{https://github.com/Lasagne/Lasagne}
Python CNN implementation to accomplish this, running on a GPU server.
A 50/50 training/testing split was used, and training run for 500
epochs on each dataset.

\subsubsection*{Detection of pathogen objects in test images}

Upon completion of CNN training, the resulting model is able to classify
a small image patch as containing an object of interest or not. The
method described above can be used to identify patches for small regions
of an image which contain the pathogens of interest. In order to identify
pathogens within the complete image or field of view, it is necessary
to split the image into patches, evaluate each one with the trained
network, and select those with high activation scores. However, this
process alone tends to identify many overlapping patches for each
actual object in the test image, particularly when a small stride
is used when creating the patches, such that there is a high degree
of overlap. For this reason, \emph{non-maximum suppression} is used
with the aim of having one activation per object within the test image.
This works by first finding overlaps amongst the selected patches
in the test image, then for those which overlap beyond a certain extent,
choosing the one with the highest probability and suppressing the
others.

\section{Results\label{sec:Results}}

\begin{figure}[t]
\begin{centering}
\begin{tabular}{>{\centering}p{0.31\textwidth}>{\centering}p{0.31\textwidth}>{\centering}p{0.31\textwidth}}
(a) Malaria & (b) Tuberculosis & (c) Hookworm\tabularnewline
\end{tabular}
\par\end{centering}

\begin{centering}
\includegraphics[width=0.33\textwidth]{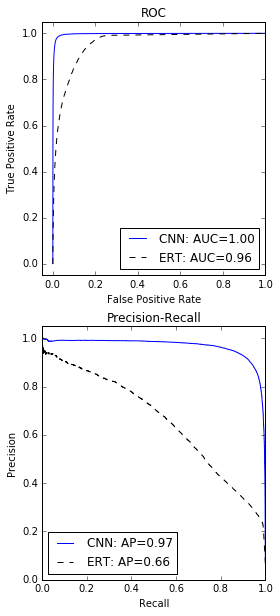}\includegraphics[width=0.33\textwidth]{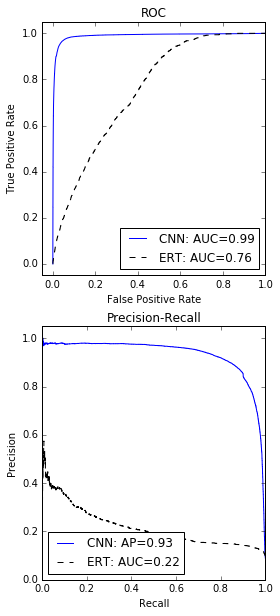}\includegraphics[width=0.33\textwidth]{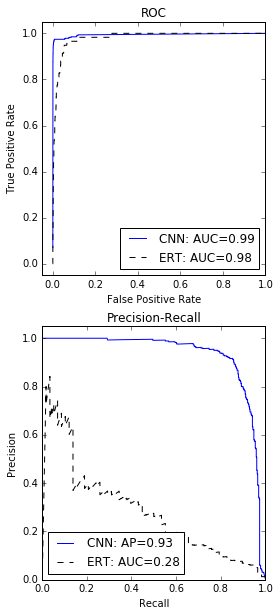}
\par\end{centering}

\caption{ROC and precision-recall for malaria, tuberculosis and intestinal
parasites detection, showing Area Under Curve (AUC) and Average Precision
(AP).}
\label{fig:roc-pr-curves}
\end{figure}

The trained networks were each applied to the corresponding test sets:
for plasmodium detection, containing 261,345 test patches (11.3\%
positive), for tuberculosis containing 315,142 test patches (9.0\%
positive), and for hookworm containing 253,503 patches (0.4\% positive).
Receiver Operating Characteristics and Precision-Recall curves are
shown for each case in Fig. \ref{fig:roc-pr-curves}. We compare this
to a more traditional method of computer vision used in other diagnosis
tasks, where shape features are extracted from each patch and applied
to a classifier. The shape features we use are a set of morphological
and moment statistics following the methodology described in \citep{19},
and applied to an Extremely Randomized Trees classifier \citep{2006extratrees}
with 100 trees. The CNN provides a dramatic increase in performance
in all cases.

\begin{figure}[t]
\includegraphics[width=1\textwidth]{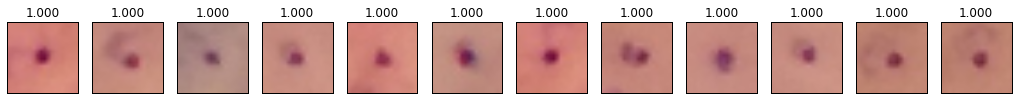}

\includegraphics[width=1\textwidth]{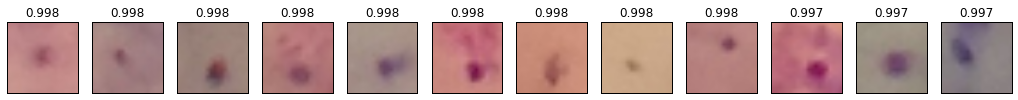}

\includegraphics[width=1\textwidth]{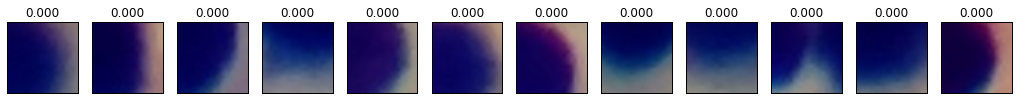}

\caption{{\small{}Plasmodium detection results, the numbers on top of each
patch denoting detection probabilities. Top row: Highest scored test
patches, all of which contain plasmodium. Middle row: the highest
scoring negative-labelled test patches, i.e. false positives. Bottom
row: lowest scoring test patches.}}
\label{fig:plasmodium-detections}
\end{figure}

\begin{figure}[t]
\includegraphics[width=1\textwidth]{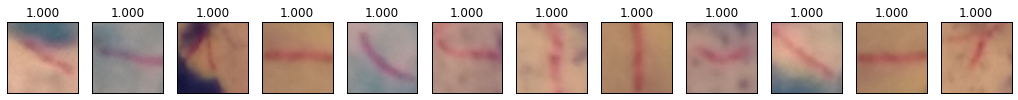}

\includegraphics[width=1\textwidth]{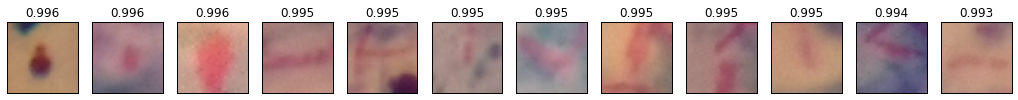}

\includegraphics[width=1\textwidth]{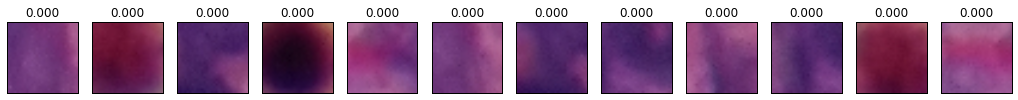}

\caption{{\small{}Tuberculosis bacilli detection results. Top row: Highest
scored test patches, all of which contain bacilli. Middle row: the
highest scoring negative-labelled test patches, many of which do contain
bacilli, and thus highlight annotation errors. Bottom row: lowest
scoring test patches. }}
\label{fig:tb-detections}
\end{figure}

The malaria detection task was that with the highest accuracy---likely
due to having the largest training set. The performance is in line with previous experimental results for malaria detection with deep learning, using images from dedicated microscope cameras rather than phones \citep{sanchez2016malaria}. Looking at the false detections
in Fig. \ref{fig:plasmodium-detections} (middle row), while these
patches were labelled as negatives, some are in fact ambiguous: it
is possible that plasmodium is present in some, but only the chromatin
dot is visible. For tuberculosis detection, we see a similar result
in Fig. \ref{fig:tb-detections} (middle row): 10 of the top 12 highest-scoring
patches labelled as negative do in fact contain bacilli, and the CNN
is therefore identifying human annotation errors. For intestinal parasites,
we used the annotated hookworm eggs as the class to detect, evaluating
the ability of the CNN to distinguish between the other types of eggs
present in the test data.

\begin{figure}

\begin{centering}
\includegraphics[width=0.45\textwidth]{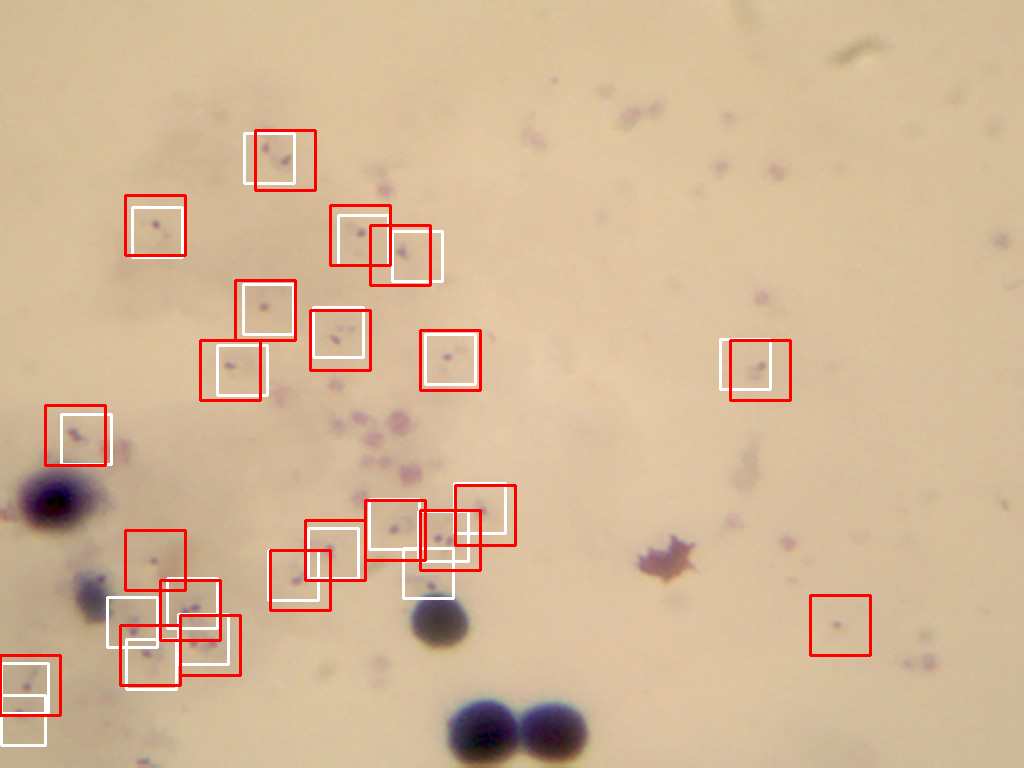}
\hspace{0.3cm}
\includegraphics[width=0.45\textwidth]{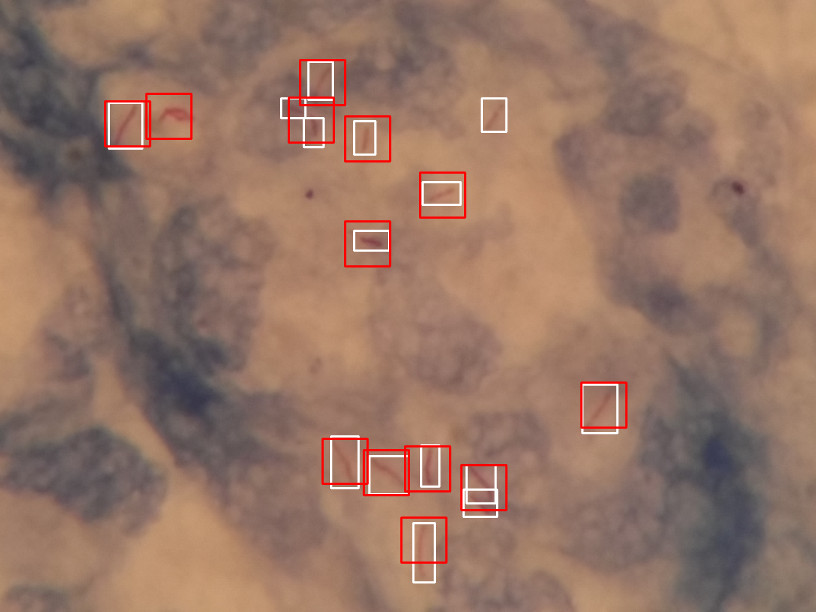}
\caption{Detected objects in test images, for plasmodium (left) and tuberculosis bacilli (right). White boxes show expert annotations on pathogen locations. Red boxes show detections by the system.}
\label{fig:detectedobjects}
\end{centering}

\end{figure}

Fig. \ref{fig:detectedobjects} shows detection on two entire test
images, when split into patches, assessing CNN activation for each
patch, and then applying non-maximum suppression. The locations of
detected objects can be seen to closely correspond to the human expert
annotations.

\begin{figure}[t]
\includegraphics[width=1\textwidth]{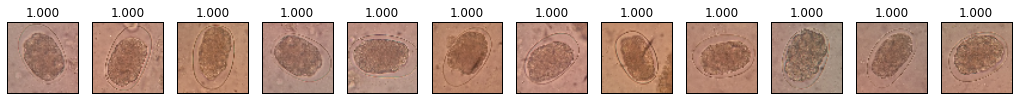}

\includegraphics[width=1\textwidth]{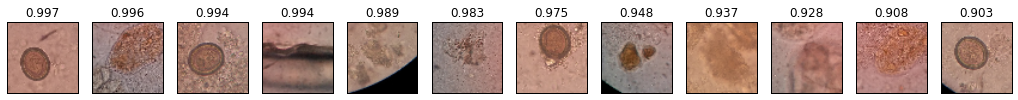}

\includegraphics[width=1\textwidth]{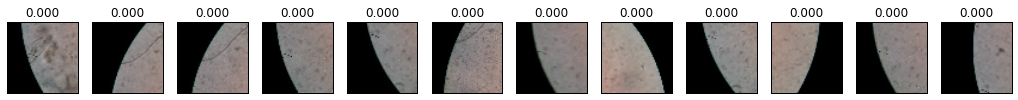}\label{fig:hookworm-detections}

\caption{{\small{}Hookworm egg detection results. Top row: Highest scored test
patches, all of which contain hookworm eggs. Middle row: the highest
scoring negative-labelled patches, i.e. false positives, mainly containing
either eggs of another parasite (Taenia) or fecal impurities. Bottom
row: lowest scoring test patches.}}
\end{figure}

\section{Discussion and Related Work\label{sec:Discussion-and-Related}}

Microscopy image analysis has been a long-standing area of research for the improvement of pathology detection and . Deep convolutional networks in particular have been used previously for detection of malaria \citep{sanchez2016malaria} and intestinal parasites \citep{peixinho2015diagnosis}, on images from conventional microscope cameras. Deep learning has been used in \cite{13} to detect the analysible metaphase chromosomes for laboratory diagnosis of leukemia, with an eight-layer multilayer perceptron network. Fully Convolutional Regression Networks (FCRNs) have been used in \cite{24} for automatic cell counting in microscopy images.
Due to a lack of labelled microscopy labelled images, CNNs application has also been extended to classification and segmentation using  multiple instance learning, which enables training CNNs using full resolution microscopy images with global labels \citep{23}. 


 Malaria diagnosis methods are reviewed in \cite{2016malariadetectionreview} with the
conclusion that improvements in accuracy are still needed. All the
methods reviewed rely on hand-tuned feature extraction, for example
using thresholding or segmentation based on hue, followed by classification
either with hand-coded decision rules or classifiers such as kNN or
SVM. Work on automated tuberculosis detection has followed a similar
pattern, generally using shape or colour features followed by application
of a classification algorithm, for example SVM classification \citep{4}
and multilayer perceptron networks \citep{9}. In \cite{12}, connected
component labeling, size thresholding, proximity grouping and an area
based classification was implemented on 205 images. Previous studies
on automated helminth detection likewise employ techniques such as
multilayer perceptrons \citep{2001helminthANN} and SVM \citep{2009helminthSVM}. 

In this paper we have shown that the performance improvement by deep
learning and CNNs compared to alternatives in other application domains
can successfully be translated to point of care microscopy-based laboratory
diagnosis. In contrast to systems with hand-engineered features for
each problem, in this case the method learns good representations
of data directly from the pixel data. The fact that in our experiments
the same network architecture successfully identifies objects in three
different types of sample further indicates its flexibility; better
results still are likely given task-specific tuning of model parameters
with cross validation for each case. This improvement in performance
can advance microscopy-based POC diagnostics which is particularly
relevant in the developing world where both microscopes and smartphones
are more readily available than skilled laboratory staff. Even where
laboratory staff are present in this context, this type of system
can be utilised as a decision support tool, identifying possible pathogens
in an image, with the technician making the final decision. This mode
of use can help laboratory staff to achieve consistency in diagnosis,
and by focusing concentration on parts of the images likely to contain
pathogens, may also help to relieve operator fatigue and improve throughput
rates.

\subsection*{Acknowledgments}
The work was funded by Grand Challenges Canada, under the Stars in Global Health program. Image annotation was carried out by Vincent Wadda, David Byansi and Patrick Byanyima, and Ezra Rwakazooba assisted with software development. We thank the anonymous reviewers whose comments helped to improve the paper.

{\small{}\bibliographystyle{plain}
\bibliography{sample,mlhc2016}
}{\small \par}
\end{document}